\documentclass[fleqn,10pt]{wlscirep}
\usepackage[utf8]{inputenc}
\usepackage[T1]{fontenc}
\usepackage{soul}
\usepackage{multirow}
\usepackage{CJKutf8}
\newcommand{\zh}[1]{\begin{CJK}{UTF8}{gbsn}#1\end{CJK}}
\usepackage{graphicx}
\usepackage{tabularx}
\usepackage{subfigure}
\usepackage{amsmath}
\usepackage{arydshln}
\usepackage[normalem]{ulem}
\usepackage{array}
\newcolumntype{C}[1]{>{\centering\arraybackslash}m{#1}}
\useunder{\uline}{\ul}{}
\title{An Evaluation of Estimative Uncertainty in Large Language Models}

\author[1]{Zhisheng Tang}
\author[1]{Ke Shen}
\author[1,*]{Mayank Kejriwal}

\affil[1]{University of Southern California, Information Sciences Institute, Marina del Rey, 90292, United States of America}

\affil[*]{kejriwal@isi.edu}

\begin{abstract}

Words of estimative probability (WEPs), such as ``maybe'' or ``probably not'' are ubiquitous in natural language for communicating estimative uncertainty, compared with direct statements involving numerical probability. Human estimative uncertainty, and its calibration with numerical estimates, has long been an area of study -- including by intelligence agencies like the CIA. This study compares estimative uncertainty in commonly used large language models (LLMs) like GPT-4 and ERNIE-4  to that of humans, and to each other. Here we show that LLMs like GPT-3.5 and GPT-4 align with human estimates for some, but not all, WEPs presented in English. Divergence is also observed when the LLM is presented with gendered roles and Chinese contexts. Further study shows that an advanced LLM like GPT-4 can consistently map between statistical and estimative uncertainty, but a significant performance gap remains. The results contribute to a growing body of research on human-LLM alignment.

\end{abstract}
\begin{document}

\flushbottom
\maketitle

\thispagestyle{empty}

\section*{Main}

In natural language, commonsense expressions of uncertainty play an important role in human communication,\cite{erev1990verbal, juanchich2020people} allowing people to account for the uncertain nature of the world without always having to rely on numbers and statistics. For example, given that it is been rainy for the past three days, a statement like ``it is highly likely that tomorrow will also be rainy'' is more common and natural than ``I estimate that the probability that it will rain tomorrow is 95 percent.'' Such words or phrases (``highly likely'') are known as \textit{Words of Estimative Probability} (WEPs) and intended to express estimative uncertainty in natural language. In human language, these words play a critical role in nuanced discussions. They allow us to express beliefs, make predictions, and communicate uncertainties in a way that can be understood and acted upon by others. WEPs make conversations more nuanced, allowing us not just to express uncertainty, but convey politeness and foster credibility. \cite{hyland1996writing, vlasyan2018linguistic, friedman2012communicating} 

First investigated systematically by Sherman Kent\cite{kent1964words} during his time at the Central Intelligence Agency (CIA), WEPs have also been studied by intelligence units and geopolitical experts owing to their ubiquity in reports, and their importance as linguistic elements in conveying varying degrees of uncertainty and ambiguity. Kent specifically sought to quantify the numerical probability (which can be a distribution) that people implicitly refer to when using different WEPs.\cite{kent1964words} Using carefully constructed surveys, he was able to map key WEPs into probability distributions that then came to be used by the CIA. Numerous works have since sought to understand the meaning and use of WEPs. \cite{beyth1982probable, friedman2015handling, shinagare2019radiologist, wallsten1986measuring, lenhardt2020likely} In his handbook, Barclay\cite{barclay1977handbook} references a survey among the NATO officers on the associated numerical probabilities for different WEPs. More recently, Fagen-Ulmschneider\cite{fagen2019perception} surveyed 123 people on their perception of probabilistic words via social media, and found that current perceptions of these WEPs have remained largely consistent with those found in Kent's earlier study. 


Recent advent, and continued application, of Large Language Models \cite{zhao2023survey} (LLMs) introduces a new dimension to this research. LLMs are pre-trained on a vast amount of text data, which includes books, articles, and websites, enabling them to grasp a wide range of linguistic patterns, cultural nuances, and factual information. As a result, they demonstrate humanlike performance on many natural language processing tasks.\cite{achiam2023gpt, touvron2023llama} Because of their generative and conversational capabilities, they are already being widely adopted in commercial applications that people interact with on a daily basis, including conversational assistants,\cite{GPT-3.5-turbo, GPT-4} automated summarization,\cite{zhang2024benchmarking} and customer service.\cite{LLM_cus} Given this adoption, the ability of LLMs to correctly use and understand WEPs is going to be critical for effective collaboration between humans and LLM-powered systems, including establishing human-AI alignment.\cite{HAIA1,HAIA2,HAIA3,HAIA4} 

This study investigates whether LLMs can express uncertainties in a way that closely mirrors human practices. This examination is motivated by the observation that how LLMs interpret WEPs may impact the trust and reliability people place in them during use. 
We begin by baselining estimative uncertainty in several important LLMs to distributional data constructed from (externally conducted) human surveys as a reference. To ensure robustness, we evaluate LLMs' estimative uncertainty under different controls, including whether \textit{ gender} and the choice of \textit{language} (English or Chinese) make a difference to the baseline findings. For example, we consider whether adding a gendered role to the prompt that is presented to an LLM affects any of the conclusions. In a similar vein, we also investigate changes in effect strength and direction both when an LLM like GPT-4 (that understands both English and Chinese) is prompted using Chinese, as well as when the LLM (such as ERNIE-4\cite{ERNIE}) is trained primarily using Chinese text. The latter experiment is motivated by the fact that LLMs are increasingly being used for multi-lingual tasks like machine translation.\cite{LLM-MT1,LLM-MT2} To investigate these effects systematically, we construct datasets involving different scenarios, controls, and WEPs. A suite of LLMs (GPT-3.5-Turbo\cite{GPT-3.5-turbo}, GPT-4\cite{GPT-4}, Llama-series models e.g., Llama-2-7B\cite{touvron2023llama}, Llama-2-13B\cite{touvron2023llama} and ERNIE-4\cite{ERNIE}) are examined.

Our second objective considers an issue that is especially important for communicating scientific information in everyday language. In scientific discourse, \textit{statistical} uncertainty tends to dominate when presenting findings, especially in formal publications. In contrast, estimative uncertainty is the norm for communicating uncertainty in everyday natural language. LLMs are starting to be used increasingly often in science, including for summarizing scientific findings.\cite{LLM-sci1,LLM-sci2} Therefore, for a specific high-performing LLM (GPT-4), we consider whether different levels of \textit{statistical} uncertainty in the prompt, appropriately controlled, lead to consistent changes (or lack thereof) in the LLM's elicitation of \textit{estimative} uncertainty. Because formal evaluation of such consistency in AI systems has not been explored thus far in the literature, we propose and formalize four novel \textit{consistency metrics} (see \textit{Methods} for details) for evaluating the extent to which an LLM like GPT-4 is able to \textit{change} its level of estimative uncertainty when prompted with changing levels of statistical uncertainty.  

Experimentally, we find that most LLMs, especially the GPT models, closely align with human uncertainty estimates in contexts of extreme or balanced certainty, and reflect a high degree of sophistication in interpreting WEPs that express positive certainty. However, we also found that of the 17 WEPs we consider, the probability estimates from GPT-3.5 and GPT-4 diverge from those of humans for 11 and 12 WEPs, respectively. Hence, alignment between humans and LLMs on a significant portion of WEPs is still wanting.  

When narratives involve gender-specific contexts, all LLMs displayed a higher divergence from human estimations, suggesting a gap in processing gender-centered context. GPT-3.5 stood out for its lower divergence from human judgments across various contexts, even when compared with GPT-4, an LLM with more advanced natural language understanding capabilities. Intriguingly, LLMs' uncertainty estimates showed only minor differences when prompted using either English or Chinese. However, ERNIE-4, an LLM pre-trained using mostly Chinese text, shows significantly different uncertainty estimates for several WEPs compared to those of GPT-3.5, using the same Chinese prompts. Because GPT-3.5 can also be prompted in Chinese, the divergence cannot be explained away simply due to the latter's lack of Chinese understanding.  

Finally, our investigation into LLMs' performance on statistically uncertain data underscored the effectiveness of GPT-4, which yielded better-than-random performance across all four consistency metrics. However, using more advanced prompting methods like Chain-of-Thought (CoT) prompting \cite{wei2022chain} failed to realize the improvements in performance that have been observed for more traditional natural language understanding problems, like question answering and commonsense reasoning. \cite{wang2022self, wei2022chain} The findings suggest that further bridging the gap between statistical uncertainty and estimative uncertainty in LLMs may prove to be more challenging using currently popular prompt engineering methods.

\section*{Results}\label{results}

\subsection*{Comparing estimative uncertainty in LLMs to humans under different experimental conditions}

\begin{figure}[h!]
    \centering
    \includegraphics[width=\textwidth]{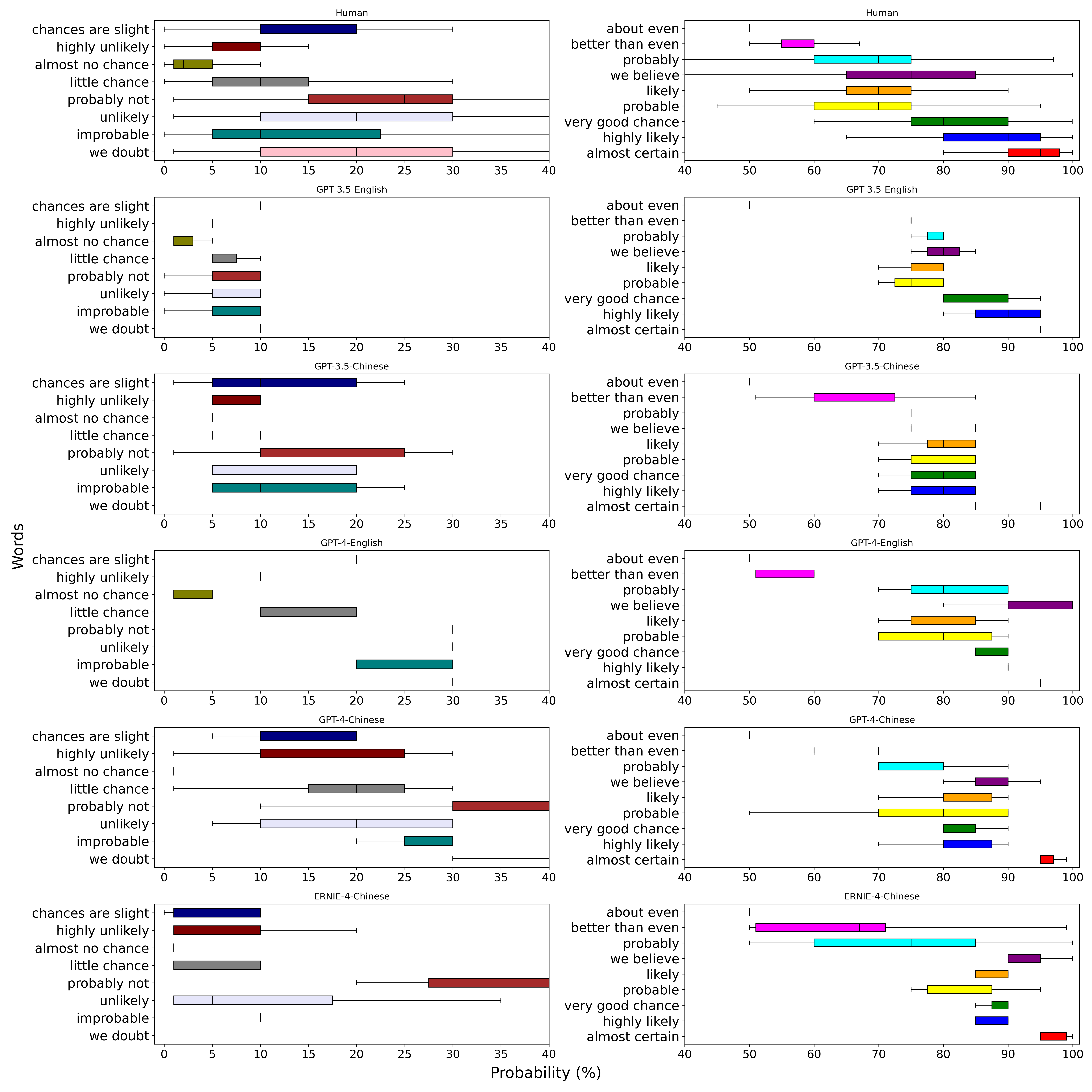}
    \caption{Distributions of probabilities (expressed as percentages on the x-axis) on 17 words of estimative probability (WEPs) elicited from six sources: \textit{human, GPT-3.5-English, GPT-3.5-Chinese, GPT-4-English, GPT-4-Chinese,} and \textit{ERNIE-4-Chinese}. The graphs on the left feature an x-axis range of 0 to 40 and include 8 WEPs on the y-axis, while the graphs on the right have an x-axis range of 40 to 100 and present the other 9 words on the y-axis. Outliers are omitted from the box-and-whisker plots, and there is zero variability in the cases where only - is indicated.}
    \label{fig:human_vs_gpt}
\end{figure}

\begin{figure}[h!]
    \centering
    \includegraphics[width=\textwidth]{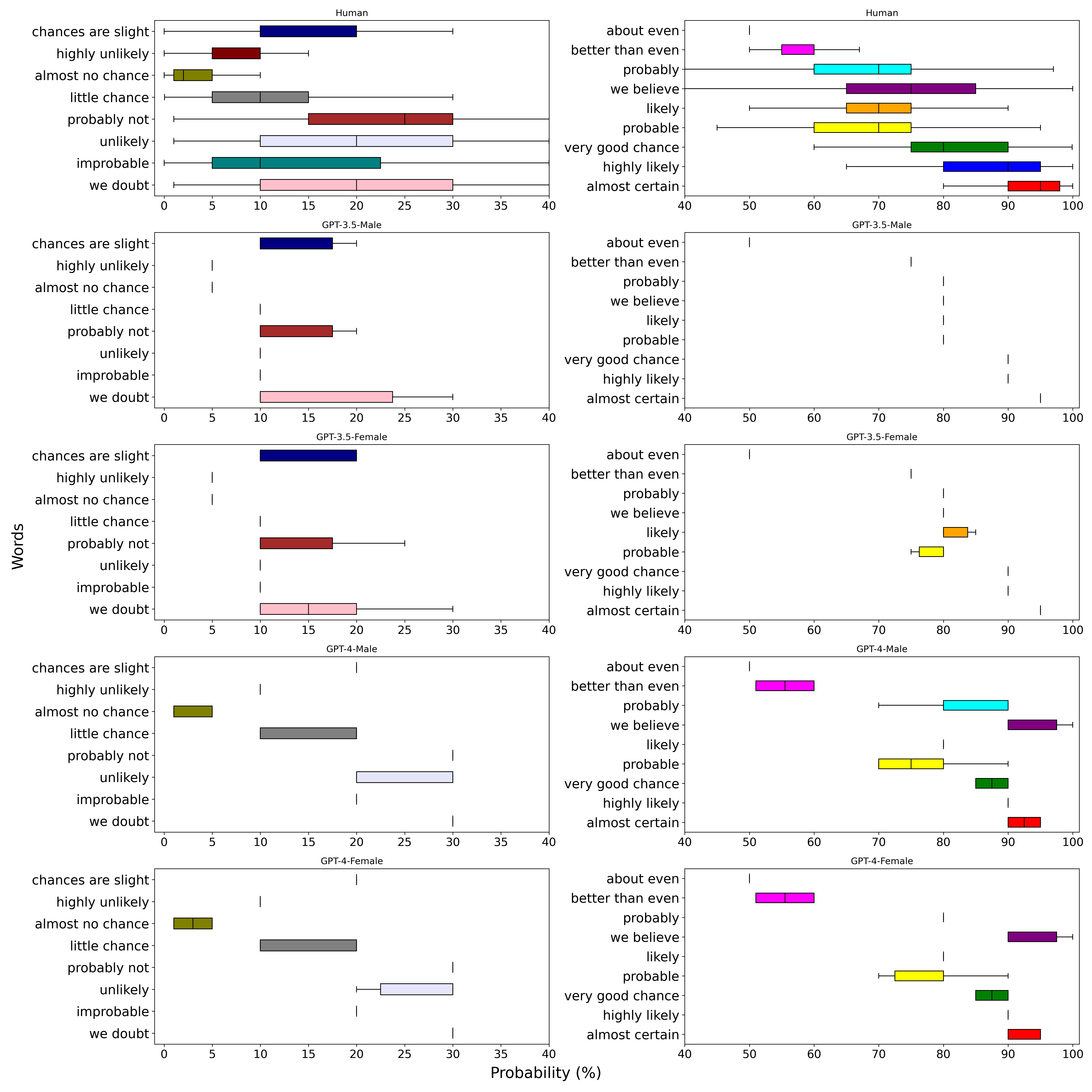}
    \caption{Distributions of probabilities (expressed as percentages on the x-axis) on 17 words of estimative probability (WEPs) elicited from five sources: \textit{human, GPT-3.5-Male, GPT-3.5-Female, GPT-4-Male,} and \textit{GPT-4-Female}. The graphs on the left feature an x-axis range of 0 to 40 and include 8 WEPs on the y-axis, while the graphs on the right have an x-axis range of 40 to 100 and present the other 9 words on the y-axis. Outliers are omitted from the box-and-whisker plots, and there is zero variability in the cases where only - is indicated.}
    
    \label{fig:human_vs_gpt_gender}
\end{figure}

\begin{figure}[h!]
    \centering
    \includegraphics[width=\textwidth]{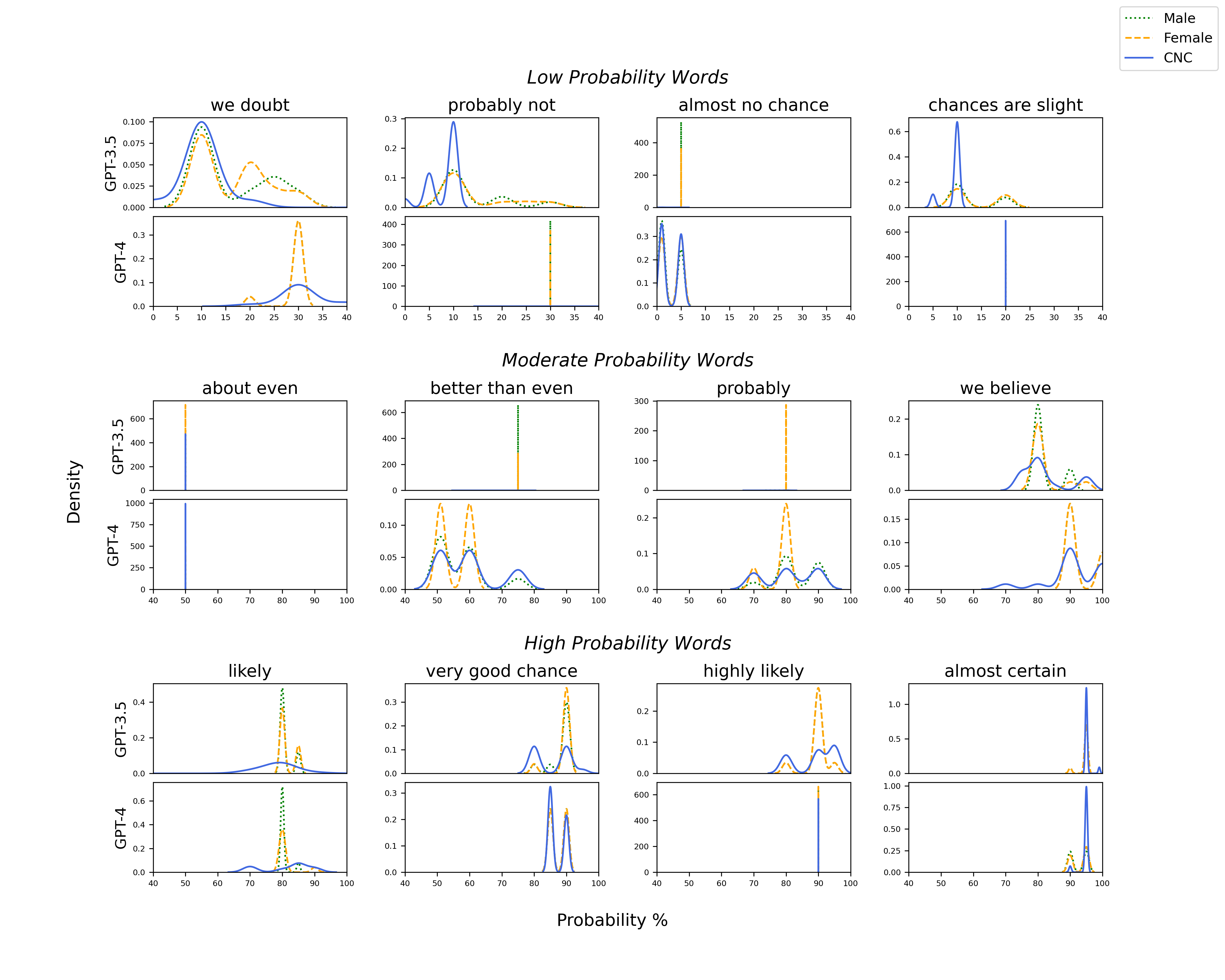}
    \caption{Distributions of probability estimations on 12 WEPs (divided into three categories: low, moderate, high probability WEPs) by GPT-3.5 and 4. Each graph shows the estimations given under \textit{Male, Female,} and \textit{Concise Narrative Context (CNC)} context settings. The last of these is gender-neutral and serves as a reference. The graphs with low probability words feature an x-axis range of 0 to 40, while the other graphs have an x-axis range of 40 to 100.}\label{fig:gender_smooth}
\end{figure}

\begin{figure}[h!]
    \centering
    \includegraphics[width=\textwidth]{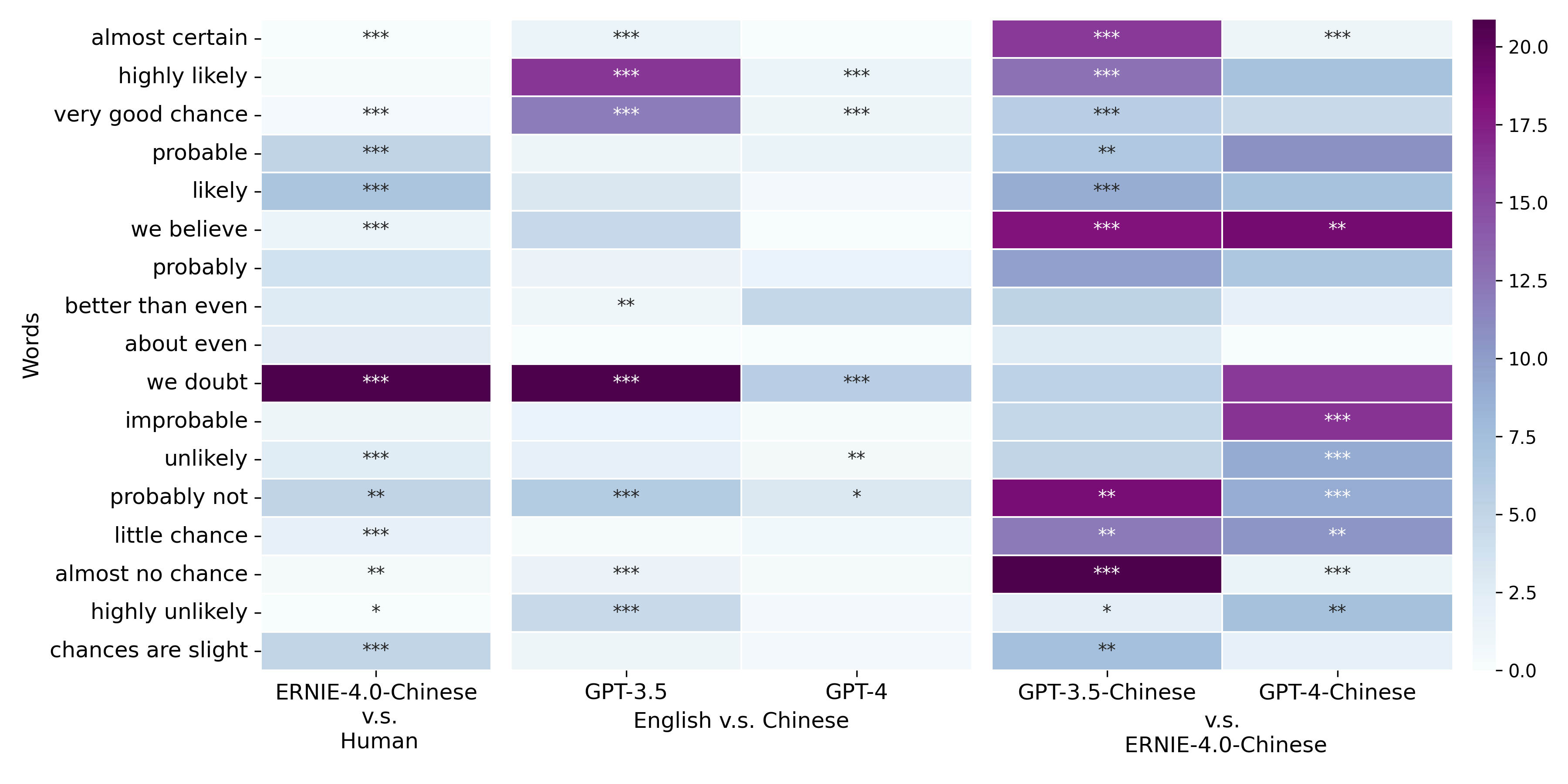}
    \caption{A heat map displaying the Kullback-Leibler (KL) divergence between various comparison pairs on 17 words of estimative probability. These pairs are (1) ERNIE-4.0 (prompted in Chinese) and humans, (2) GPT-3.5 or GPT-4 which is prompted using English and Chinese, (3) GPT-3.5 or GPT-4 compared with ERNIE-4.0 (all prompted in Chinese). The intensity of the color within each cell corresponds to the KL divergence values, with darker colors indicating higher divergence. *, **, and *** represent significant statistical significance for the Mann-Whitney U test at confidence levels of 90\%, 95\%, and 99\% levels, respectively. Supplementary Information Figures S19-S21 contain the precise Kolmogorov–Smirnov (KS) statistics used to assess the significance of these divergences.
    }
    \label{fig:chinese_sign}
\end{figure}

Figure \ref{fig:human_vs_gpt} shows the distribution of probability estimates for 17 Words of Estimative Probability (WEPs) provided by GPT-3.5 and GPT-4, aggregated across independent concise contexts presented in English and Chinese. It also includes results from ERNIE-4.0, an LLM pre-trained primarily on Chinese text, which is prompted using only Chinese. The results show that the distributions for GPT-3.5 and GPT-4 diverge from those of humans for 11 and 12 WEPs, respectively. Using the Mann-Whitney U test, the differences are found to be statistically significant. For example, there is an absolute median difference (AMD) of 5\% between the human and GPT-3.5 for the WEP `probable' ($U(N_1=123, N_2=15)=506.5, p<0.01$), with rank-biserial correlation (RBC) of $0.45$, ($95\% CI$ $0.3$ to $0.58$). There is an even larger AMD of 10\% between humans and GPT-4 ($U(N_1=123, N_2=15)=417.5, p<0.01$), with RBC of $0.49$, ($95\% CI$ $0.34$ to $0.61$). Median differences between humans and GPT-4 are also observed for WEPs such as `likely' ($AMD = 15$, $U(N_1=123, N_2=15)=407.5, p<0.01$), `we doubt' ($AMD=10$, $U(N_1=123, N_2=15)=489, p<0.01$), `unlikely' ($AMD=10$, $U(N_1=123, N_2=15)=468, p<0.01$), and `little chance' ($AMD=10$, $U(N_1=123, N_2=15)=602.5, p=0.02$). 


Interestingly, we find that humans and GPT models have statistically indistinguishable distributions for WEPs with high positive certainty, such as `almost certain' ($AMD=0$, $U(N_1=123, N_2=15)=953.5, p=0.83$) and `highly likely' ($AMD=0$, $U(N_1=123, N_2=15)=772.5, p=0.29$) for GPT-4. Similarly, humans and GPT models have AMDs of zero on `about even' ($U(N_1=123, N_2=15)=967.5, p=0.56$), for both GPT-3.5 and GPT-4. Overall, we find that GPT-3.5 consistently exhibits lower divergence than GPT-4 in most contextual analyses, despite GPT-4's superior performance in various natural language understanding tasks. \cite{achiam2023gpt} While the two still offer relatively close estimations, GPT-3.5's estimations are more closely aligned with human judgments across various contexts, suggesting that it interprets estimative uncertainty in a more human-like manner.


Figure \ref{fig:human_vs_gpt_gender} displays the distribution of probability estimates for 17 WEPs provided by GPT-3.5 and GPT-4 using gender-specific prompts. These prompts either have \textit{Male} (e.g., `he') or \textit{Female} (e.g., `she') as the subject. The first noticeable difference is that, under gender-specific contexts, GPT distributions exhibit less variability compared to human distributions; in several cases (e.g., `highly unlikely,' `improbable,' and `highly likely'), the GPT distributions even collapse into a single point. Figure \ref{fig:gender_smooth} also presents the distributions of probability estimates for 12 WEPs divided into three categories (high, moderate, and low probability WEPs). Detailed statistical analyses (Supplementary Information Figures S9-S15) show that, for individual LLMs, the gender of the subject does not yield significantly different estimations, except for `probably' ($AMD=0$, $U(N_1=10, N_2=10)=71, p=0.07$ for GPT-4). Additionally, we observe (Supplementary Information Figures S1-S8) that the estimations obtained from the GPT models, when prompted with gender-specific contexts, exhibit similar differences (compared to human estimations) as those observed when the models are prompted with gender-neutral concise narrative contexts. For the two GPT models, the differences between prompting using the male and gender-neural concise narrative context are most significant in GPT-3.5 for WEPs expressing negative certainty, such as `almost no chance' ($AMD=4$, $U(N_1=10, N_2=15)=130, p<0.01$), `little chance' ($AMD=5$, $U(N_1=10, N_2=15)=117, p=0.01$).


Finally, Figure \ref{fig:chinese_sign} presents the divergence between the probability distributions of the different models, depending on whether the prompts are in English or Chinese. On the left, it compares the responses generated by ERNIE-4.0 to Chinese prompts with those provided by humans. In the middle, it compares responses when prompted in both English and Chinese for GPT-3.5 and GPT-4. On the right, it contrasts the results from GPT-3.5 or GPT-4 with those from ERNIE-4.0, with all prompts in Chinese. Focusing on the difference between the estimations from ERNIE-4 and humans, we observe that the Kullback-Leibler (KL) divergence is low for 16 WEPs, as the color indicates, with the sole exception being `we doubt' ($AMD=40$, $U(N_1=15, N_2=123)=1778, p<0.01$). However, we also note that 12 WEPs exhibit statistically significant differences for the Mann-Whitney U test. This test can detect differences in their central tendencies or how their values are distributed, making it more sensitive to median differences between distributions, whereas KL divergence quantifies how much one distribution diverges from a second distribution. This suggests that while the overall `information content' of the compared distributions is similar, they still differ significantly in their median.

Turning to the comparison between prompting in English and Chinese using GPT-3.5 or GPT-4, we find a darker color and a higher KL divergence, suggesting a larger difference between the probability estimations when prompted using English and Chinese, for WEPs like `highly likely' ($AMD=10$, $U(N_1=15, N_2=15)=184, p<0.01$ for GPT-3.5, $AMD=10$, $U(N_1=15, N_2=15)=195, p<0.01$ for GPT-4), `very good chance' ($AMD=15$, $U(N_1=15, N_2=15)=163.5, p<0.01$ for GPT-3.5, $AMD=10$, $U(N_1=5, N_2=15)=187.5, p<0.01$ for GPT-4), and `we doubt' ($AMD=65$, $U(N_1=15, N_2=15)=15.5, p<0.01$ for GPT-3.5, $AMD=40$, $U(N_1=15, N_2=15)=14, p<0.01$ for GPT-4). Finally, we observe an even larger difference between the estimations from GPT-3.5 and ERNIE-4.0, as well as between GPT-4 and ERNIE-4.0, for WEPs such as `almost certain' ($AMD=11.5$, $U(N_1=15, N_2=15)=7.5, p<0.01$ for GPT-3.5, $AMD=7.5$, $U(N_1=15, N_2=15)=160.5, p<0.01$ for GPT-4), `we believe' ($AMD=10$, $U(N_1=15, N_2=15)=29, p<0.01$ for GPT-3.5, $AMD=5$, $U(N_1=15, N_2=15)=131, p=0.01$ for GPT-4), and `almost no chance' ($AMD=4$, $U(N_1=15, N_2=15)=202.5, p<0.01$ for GPT-3.5, $AMD=4$, $U(N_1=15, N_2=15)=33.5, p<0.01$ for GPT-4). 


Finally, we found that the probability estimates from Llama-2-7B and Llama-2-13B, prompted in English, are largely consistent with those found in the GPT models. However, their estimates often exhibit larger divergence from those of humans. These results are provided in Supplementary Information Figures S1-S8.

\subsection*{Investigating the effect of statistical uncertainty on GPT-4's estimative uncertainty}

\begin{figure}[h!]
    \centering
    \includegraphics[width=\textwidth]{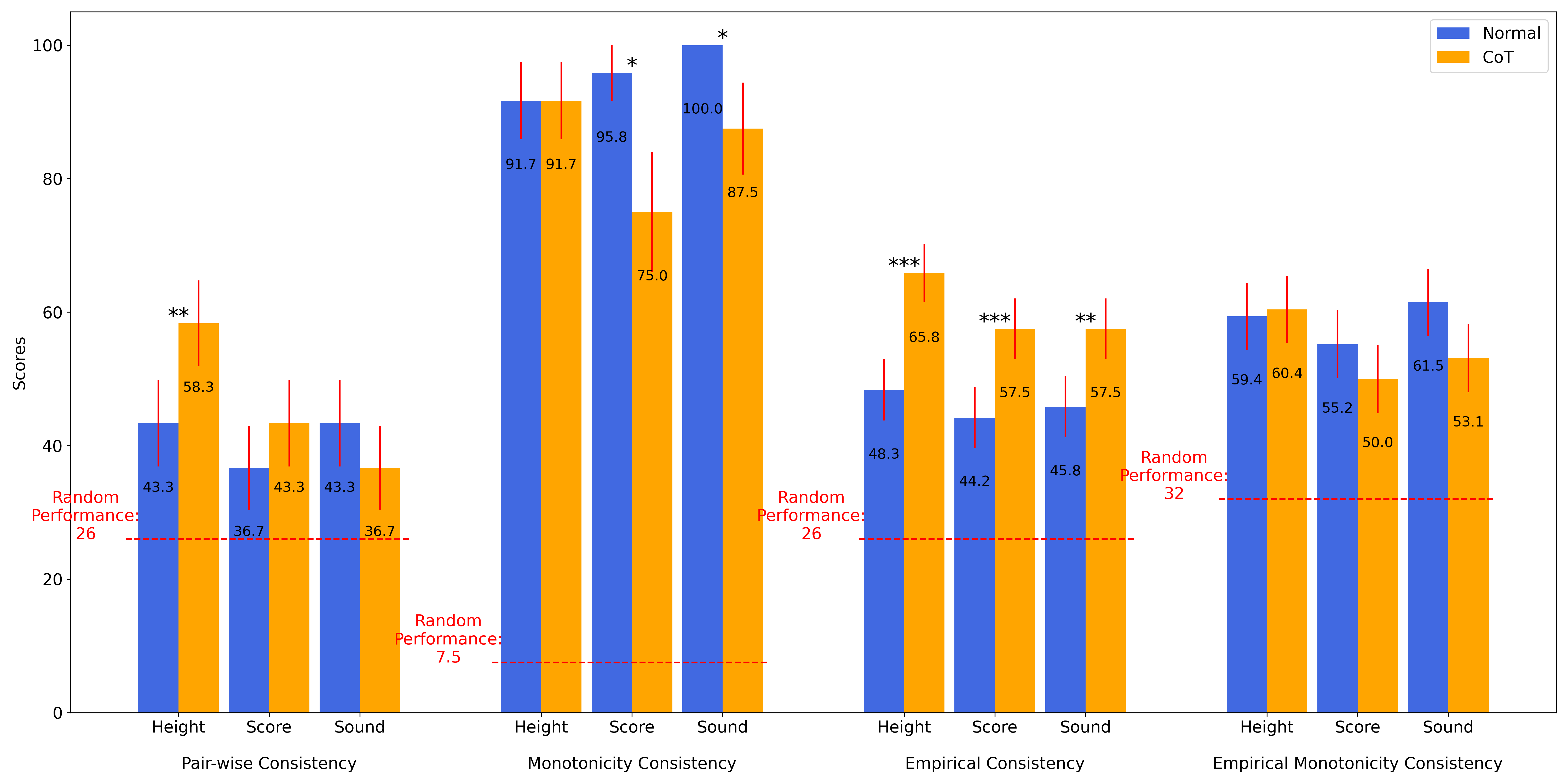}
    \caption{A bar graph illustrating the performance of GPT-4 on answering questions about the outcome of statistically uncertain events using words of estimating probability (WEPs). The graphs compare scores using four metrics: pair-wise consistency, monotonicity consistency, empirical consistency, and empirical monotonicity consistency, for both standard and Chain-Of-Thought (CoT) prompting methods. Results for each metric are further divided based on different scenarios. The random performance is shown as a red dashed line for each metric. The standard error is shown as a vertical red line, and the numerical value corresponding to each bar is displayed. *, **, and *** represent statistical significance between normal and CoT prompting, using the paired t-test, at the 90\%, 95\%, and 99\% confidence levels, respectively.}
    \label{fig:bar_graph_rq2_general}
\end{figure}

\begin{figure}[h!]
    \centering
    \includegraphics[width=\textwidth]{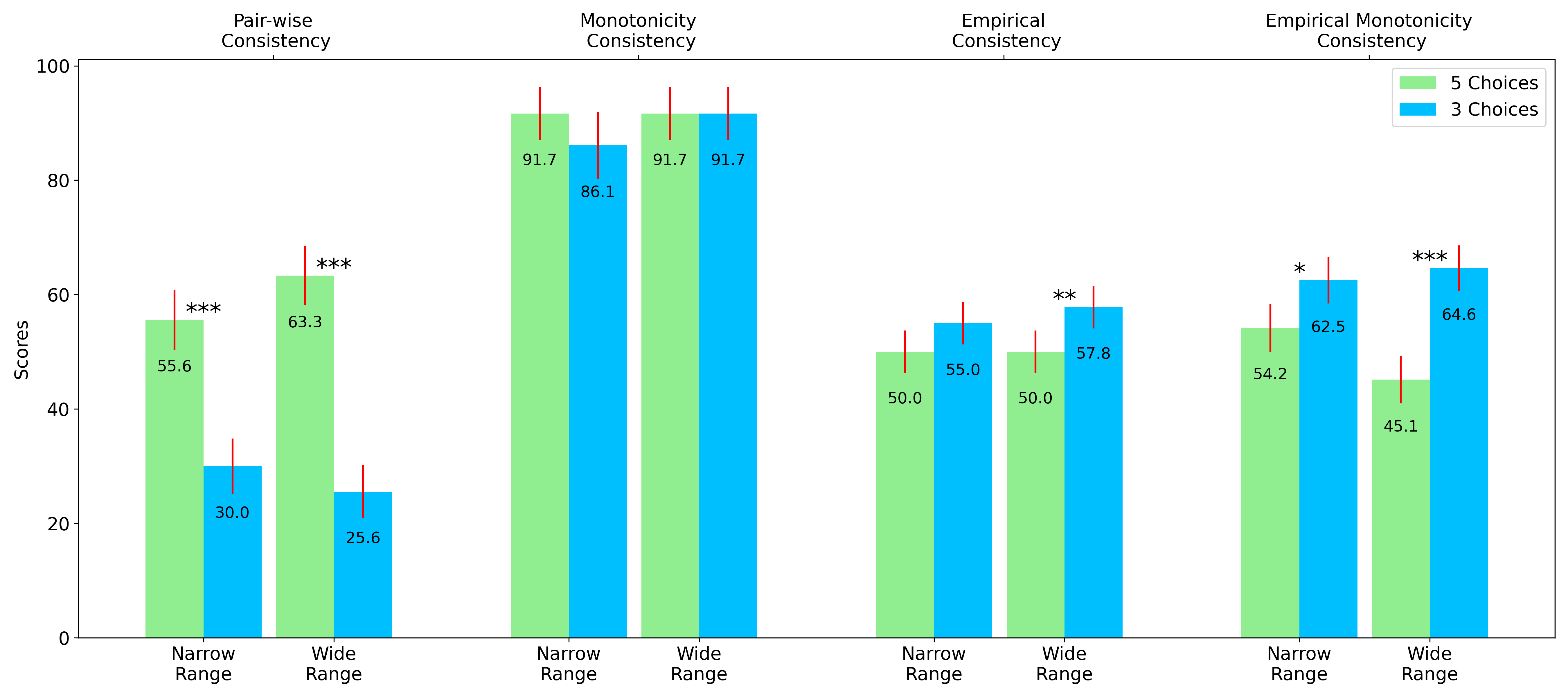}
    \caption{A bar graph comparing the performance of GPT-4 on answering questions about the outcome of statistically uncertain events using words of estimating probability (WEPs) across two settings based on the number of WEPs choices provided: one setting offers five choices, while the other offers three. The model is evaluated using four metrics: pair-wise consistency, monotonicity consistency, empirical consistency, and empirical monotonicity consistency. For each metric, we also control for the range of statistically uncertain outcomes, with a setup with a \textit{narrow} or less statistically uncertain range (left), and a \textit{wide} or more statistically uncertain range (right). The standard error is shown as a red line, and the numerical value corresponding to each bar is displayed. *, **, and *** represent  statistically significant differences between the 5-choices and 3-choices settings, using the paired t-test, at the 90\%, 95\%, and 99\% confidence levels, respectively. Supplementary Information Figure S23 also presents a similar comparison controlling for either the number of WEPs choices provided, or the range of statistically uncertain outcomes.
    }
    \label{fig:comparison_rq2}
\end{figure}

To evaluate GPT-4's performance in estimating the outcome of statistically uncertain events using WEPs, we created three different scenarios (\textit{Height, Score,} and \textit{Sound}). In general, each question in the dataset provides a set of WEP choices to the LLM, and elicits from it the choice that best describes the probability of a number falling within an interval, given a sample `distribution' of past observations. For example, one question is: \textit{Complete the following sentence using one of the choices, listed in descending order of likelihood, that best fits the sentence: A.is almost certainly B.is likely to be C.is maybe D.is unlikely to be E.is almost certainly not. I randomly picked 20 specimens from an unknown population. I recorded their heights, which are 116, 93, 94, 89, 108, 76, 117, 92, 103, 97, 114, 79, 96, 96, 111, 89, 98, 91, 100, 105. Based on this information, if I randomly pick one additional specimen from the same population, the specimen's height \_ below 99.} We elicit responses from the LLM using both standard prompting, as well as Chain-of-Thought (CoT) prompting \cite{wei2022chain} that is further detailed in \emph{Methods}.



Four metrics are proposed for evaluating the consistency of LLMs: \textit{pair-wise
consistency}, \textit{monotonicity consistency}, \textit{empirical consistency}, and \textit{empirical monotonicity consistency}. The minimum and maximum consistency score is 0 and 100, with 100 being the most consistent. However, the expected random performance for each metric is different. More details on the dataset and the metrics are provided in \emph{Methods}. 



Figure \ref{fig:bar_graph_rq2_general} displays the performance of GPT-4 evaluated on both the standard and CoT prompting methods using the four proposed metrics (\textit{pair-wise consistency, monotonicity consistency, empirical consistency, }and\textit{ empirical monotonicity consistency}). First, we observe that all the results are well above random performance, indicating the efficacy of employing LLMs in estimating probabilities from statistically uncertain data using WEPs. However, it is worth noting that these results do not achieve the same level of high performance as observed in other natural language processing or math-word tasks\cite{achiam2023gpt}. Surprisingly, the CoT prompting method \cite{kojima2022large} only gains significant performance when the LLM is evaluated using empirical consistency ($t(59)=-4.15, p<0.01$, with Cohen's $d=-0.358$ for the `Height' scenario, $t(59)=-2.82, p<0.01, d=-0.268$ for the `Score' scenario, $t(59)=-2.61, p=0.01, d=-0.234$ for the `Sound' scenario). In examining the results for monotonicity consistency, we found that the model consistently chooses the same choice for all questions instantiated using increasing confidence levels, which yields a high score but suggests a lack of nuanced understanding and calibration of uncertainty. This is confirmed by the results obtained using the empirical monotonicity consistency metric, where such a simple choice combination is not accepted, and steep performance drops are observed.

Figure \ref{fig:comparison_rq2} shows the performance of GPT-4 in two settings, based on the number of WEPs choices provided, with one setting offering \textit{5 choices} and the other \textit{3 choices}. We observe that the performance in the\textit{ 5 choices }setting is significantly higher than that on the \textit{3 choices} setting when evaluated using pair-wise consistency ($t(179)=6.48, p<0.01, d=0.673$). This might seem initially surprising because, intuitively, having fewer options should make it easier for the model to make the correct choice. However, choices in the \emph{3 choices} set may seen by the model to be less distinct from each other, making it consequently more challenging for it to perform well under this condition. However, when evaluated using empirical consistency ($t(359)=-2.35, p<0.05, d=-0.128$) and empirical monotonicity consistency ($t(287)=-4.15, p<0.01, d=-0.283$), GPT-4 does perform better under the \textit{3 choices} condition. Combined with Supplementary Information Figure S23, we observe statistically comparable performance between the narrow and wide range of the statistically uncertain outcomes for all metrics, demonstrating the robustness of GPT-4 in appropriately responding to different possible (statistically) uncertain distributions. Nonetheless, we note that the consistency is well below 100 percent on most metrics, scenarios and conditions, showing that the problem of aligning statistical uncertainty with estimative uncertainty cannot be considered to be solved, even in an advanced commercial LLM like GPT-4.


\section*{Discussion}  
In comparing uncertainty estimates between LLMs and humans, our findings show that, for 11 and 12 WEPs out of 17, GPT-3.5 and GPT-4 respectively give probability estimates that are different from those given by humans. However, in situations of high positive certainty(e.g., `almost certain' or `highly likely'), the GPT models' uncertainty estimates closely mirror those of humans. Intriguingly, GPT-3.5 consistently shows lower KL divergence when compared to GPT-4 across various scenarios despite GPT-4's advanced capabilities. When prompted under gender-specific contexts, GPT models' estimations often collapse into a single point but exhibit minimal divergence from the human estimate for the majority of WEPs. Additionally, we observed minor divergences between ERNIE-4.0's estimates and those of humans. However, a more significant divergence becomes apparent when the prompting language shifts from English to Chinese for GPT-3.5 and GPT-4. This divergence widens further when comparing the estimates from GPT-3.5 and GPT-4 with those from ERNIE-4.0, with all models being prompted in Chinese.


In investigating the capacity of LLMs to handle statistical uncertainties when using WEPs to express their estimative uncertainty, we find that the performance of GPT-4 is significantly better than random when evaluated using the four different consistency metrics, affirming the general effectiveness of LLMs in interpreting statistical uncertain data. Although GPT-4 demonstrates near-perfect performance when assessed using the monotonicity consistency metric, this appears to be an illusion, as the model consistently makes the same choices under varying conditions. This illusion signifies that GPT-4 has a limited understanding of the concepts involved. We also found that the performance of GPT-4 is sensitive to the number of choices it is allowed to choose from, but not to the range of the statistical uncertainty that is being ingested into the prompt. The little, or even negative, improvement yielded by Chain-of-Thought prompting suggests the inherent difference between this `task' and other natural language processing tasks like question answering and information extraction. \cite{wang2022self, wei2022chain, trivedi2022interleaving}

Our work builds on Kent's originally classified work \textit{Words of Estimative Probability},\cite{kent1964words} which explored human perception of probabilistic words in a systematic way. We extend the work by Fagen-Ulmschneider,\cite{fagen2019perception} who surveyed people's probability estimations on 17 different WEPs, by comparing the corresponding probability distribution of each WEP with LLMs that are pre-trained on larger amounts of human-created data. These works also inspired Sileo\cite{sileo2022probing} to investigate how challenging it is for language models to understand WEPs. However, they only use the median surveyed probability for each WEP to compare with the language models' estimations. In addition to considering the full distribution and controls such as gender and language, this study also uses more recent LLMs with more parameters and stronger conversational abilities. Furthermore, we analyzed the uncertainty estimates of GPT-4 when facing prompts with statistically uncertain data. Previous research\cite{wei2022chain, kojima2022large} showed that Chain-of-Thought prompting can be used to boost performance on numerical- or math-related natural language problems. However, our investigation reveals limitations in CoT's ability to increase alignment between statistical and estimative uncertainty compared to standard prompting. 

Our research also aligns with earlier research contributions in `BERTology,' \cite{rogers2021primer} which sought to investigate the behavior and underlying mechanisms of the BERT language model\cite{devlin2018bert} and its derivatives. This seminal work has paved the way for a new line of studies that emphasized greater understanding of language model behavior compared to simply measuring their performance on task-oriented benchmarks. Additionally, Binz and Schulz \cite{binz2023using} redirect the emphasis toward LLMs, comparing their performance with human behaviors in psychological experiments originally designed for human subjects. Our contribution adds to the growing body of research that explores the alignment between LLMs and humans\cite{wang2023aligning} and investigates such alignment in the understanding of WEPs.

A key limitation of this study is the use of human uncertainty estimates from a sample of undergraduate students obtained from a survey. While other work has pointed toward the reliability of the survey, it is not settled whether these estimates are representative of the broader population's understanding of WEPs, which can skew the comparative analysis between humans and LLMs. Future work should include a more diverse population when surveying human uncertainty estimations. Additionally, we rely on artificially generated data to evaluate GPT-4's performance on prompts with statistically uncertain data. While this paradigm offers a controlled environment to assess an LLM's capabilities, it might not be accurate enough to represent real-world statistical scenarios and the complexities involved in interpreting statistically uncertain data in a natural language context. A promising future direction is to explore these under real-world scenarios and data (e.g., using carefully selected prompts containing statistical information reported in actual scientific publications), which may yield insights with stronger external validity. Additionally, we only focus on the 17 WEPs that were used in the survey, whereas incorporating a wider range of probabilistic expressions could provide more comprehensive insights into LLMs' estimations. While we found little difference in prompting either in Chinese or in English for the GPT models, a cross-linguistic study with more languages in uncertainty estimation is merited. Moreover, our study focused on static prompts without considering the dynamic nature of conversations. Future works could examine how LLMs adjust their probability estimations in response to a changing context within a dialogue.

\section*{Methods}\label{sec:method}
The structure of this section is organized into two distinct parts, each dedicated to the two research objectives. Methods for each objective are discussed in detail from three perspectives: \textit{data construction, metrics,} and \textit{experimental setup}. In the section on data construction, we delineate the methodologies employed to curate the datasets tailored for our experiments, and the additional context for our decision to construct them in that manner. The metrics section provides a detailed explanation of both traditional and innovative criteria used to evaluate the results. Lastly, the experimental setup section provides a description of the specific LLMs used, along with other relevant details on the evaluation framework. 

\subsection*{Comparing estimative uncertainty in LLMs to humans under different experimental conditions}
\subsubsection*{Data Construction}

The first research question aims to compare the interpretation of estimative uncertainty in WEPs using numerical probabilities between LLMs and humans. To enable meaningful comparison between humans and LLMs, we utilize the same set of WEPs as used in the Fagen-Ulmschneider's survey.\cite{fagen2019perception} This set of seventeen WEPs contains \textit{almost certain, highly likely, very good chance, probable, likely, we believe, probably, better than even, about even, we doubt, improbable, unlikely, probably not, little chance, almost no chance, highly unlikely}, and \textit{chances are slight}. Additionally, we introduce four distinct context settings inspired by the narratives found in Kent's CIA report\cite{kent1964words} and an article from Harvard Business Review.\cite{HarvardBusinessReview_2018} These context settings, comprising manually crafted context-templates, are specifically designed to evaluate if the certainty estimations made by LLMs vary based on different narrative backgrounds. The settings are described in Table \ref{table: tempOverview}, which also provides a sample template and related counts and statistics for each of these contexts, with Tables \ref{sup_cnc}, \ref{sup_enc}, and \ref{sup_fcnc} providing the full list of templates for completeness:

\begin{itemize}
    \item \emph{Concise Narrative Context} (CNC): Simple, intuitive narrative contexts that include brief scenarios intended to provide a straightforward background to use WEPs.
    \item \emph{Extended Narrative Context} (ENC): In contrast to the CNC setting, ENCs contain more prolonged narratives with increased information, incorporating various clauses to create a detailed setting.
    \item \emph{Female-Centric Narrative Context} (FCNC): Similar to the CNC in its simplicity, this setting includes only short scenarios. However, it specifically employs `She' as the subject to introduce a gender-specific narrative.
    \item \emph{Male-Centric Narrative Context} (MCNC): This mirrors the FCNC in narrative structure but replaces `She' with `He' as the subject, thus providing a comparative perspective on gender-based narrative interpretation.
\end{itemize}

\begin{table}[h!]
\begin{tabular}{m{2cm}m{2.2cm}m{3.2cm}m{8cm}}
\hline
                                       & \textbf{$\#$templates} & \textbf{Average template length} & \textbf{Example template}                                                                                                                                                                                                                                         \\ \hline
\textbf{CNC}              & 15            & 7.1                  & They will \{\} launch before us.                                                                                                                                                                                                                   \\ \hline
\textbf{ENC}            & 11            & 24.3                 & Given the diverse sources of the intelligence report, it is \{\} a mistake that this piece of information was overlooked, though there are indications that it could have been due to a human error.\\ \hline
\textbf{FCNC/MCNC} & 10            & 8.6                  & She / He \{\} orders the same dish at that restaurant.                                                                                                                                                                                             \\ \hline
\end{tabular}
\caption{An overview of four context settings that are compatible with different WEPs. Note that \{\} in \emph{Example} columns represents a placeholder for any WEP. Tables \ref{sup_cnc}, \ref{sup_enc}, and \ref{sup_fcnc} provide the full list.}\label{table: tempOverview}
\end{table}

\begin{table}[h]
    \centering
    \begin{tabular}{l}
    \hline
    The film festival \{\} attracts a large audience. \\
    They will \{\} launch before us. \\
    The local concert \{\} sells out quickly. \\
    The charity gala \{\} raises significant funds. \\
    The art exhibition \{\} receives positive reviews. \\
    That antique fair \{\} unveils rare collectibles. \\
    The mountain trail \{\} offers breathtaking views at dawn. \\
    The computer \{\} malfunctions when I have an important task to complete. \\
    The museum \{\} gets crowded on weekends. \\
    They are \{\} moving to Spain for the summer. \\
    It is \{\} a military airfield. \\
    The restaurant is \{\} the cheapest option available. \\
    The theory is \{\} the only explanation for the phenomenon. \\
    This ingredient is \{\} necessary for the recipe. \\
    Based on the weather forecast, it will \{\} rain tomorrow. \\
    \hline
    \end{tabular}
    \caption{The full list of \textit{Concise Narrative Context} (CNC) templates. \{\} represents a placeholder for possible WEPs.}\label{sup_cnc}
\end{table}

\begin{table}[h]
\centering
\begin{tabular}{p{\textwidth}}
\hline
Khrushchev may have had in the back of his mind such and such, or indeed it is \{\} that somebody had just primed him with a particular perspective or piece of information that influenced his decision-making at that moment. \\
\hline
It's \{\} that when faced with the crisis, Churchill recalled past failures, or it's conceivable that an advisor had recently presented him with fresh insights that swayed his judgment. \\
\hline
In his diplomatic endeavors, Ahmed \{\} held the lessons from his predecessors in high esteem. \\
\hline
Given the intricate nature of the puzzle, solving it in under an hour is \{\} a remarkable feat. \\
\hline
In the realm of popular music, where artists come and go with the changing trends and fans chase the latest hits, crafting a timeless song that resonates with multiple generations is \{\} an achievement signifying true artistry. \\
\hline
Given the diverse sources of the intelligence report, it is \{\} a mistake that this piece of information was overlooked, though there are indications that it could have been due to a human error. \\
\hline
While the painting is \{\} from the Renaissance period, it sometimes carries motifs typical of that era; artists always borrow inspiration from the past. \\
\hline
The intricate web of conspiracy theories surrounding the moon landing suggests that it was \{\} a hoax perpetuated by NASA \\
\hline
Despite the complexity of climate models, they indicate that global temperatures will \{\} decrease significantly in the coming decades. \\
\hline
The historical evidence suggests that it was \{\} a coincidence that these two great inventors were born in the same era. \\
\hline
The chances of winning the lottery are \{\} in your favor, but that doesn't stop millions of people from trying their luck. \\
\hline
\end{tabular}
\caption{The full list of \textit{Extended Narrative Context} (ENC) templates. \{\} represents a placeholder for possible WEPs.}\label{sup_enc}
\end{table}

\begin{table}[h]
\centering
\begin{tabular}{l}
\hline
She / He \{\} wakes up at 6 a.m. \\
She / He \{\} takes the bus to work. \\
She / He \{\} orders the same dish at that restaurant. \\
She / He \{\} attends the weekly meetings. \\
She / He \{\} visits the park on weekends. \\
She / He \{\} reads a book before bed. \\
She / He \{\} remembers to bring an umbrella when it's cloudy. \\
She / He \{\} dines out on Fridays. \\
She / He \{\} listens to the news on the morning drive. \\
She / He \{\} bakes a cake for birthdays. \\
\hline
\end{tabular}
\caption{The full list of \textit{Female-Centric Narrative Context} (FCNC) and \textit{Male-Centric Narrative Context} (MCNC) templates. ``/'' represent a choice between ``She" and ``He''. The FCNC templates choose ``She" as the subject and the MCNC templates choose ``He'' as the subject. \{\} represents a placeholder for possible WEPs.}\label{sup_fcnc}
\end{table}

\begin{table}[h]
    \centering
    \resizebox{\columnwidth}{!}{
    \begin{tabular}{l|l}
    \hline
    English & Chinese \\
    \hline
    The film festival \{\} attracts a large audience. & \zh{电影节\{\}吸引大量观众}\\
    They will \{\} launch before us. & \zh{他们\{\}在我们之前发布}\\
    The local concert \{\} sells out quickly. & \zh{当地音乐会\{\}很快售罄} \\
    The charity gala \{\} raises significant funds. & \zh{慈善晚会\{\}筹集到大量资金}\\
    The art exhibition \{\} receives positive reviews. & \zh{艺术展览\{\}收到积极评价}\\
    That antique fair \{\} unveils rare collectibles. & \zh{古董展\{\}展示稀有收藏品}\\
    The mountain trail \{\} offers breathtaking views at dawn. & \zh{山道\{\}在黎明时分提供令人叹为观止的景色}\\
    The computer \{\} malfunctions when I have an important task to complete. & \zh{当我有重要任务要完成时，计算机\{\}出现故障}\\
    The museum \{\} gets crowded on weekends. & \zh{博物馆\{\}在周末拥挤} \\
    They are \{\} moving to Spain for the summer. & \zh{他们\{\}去西班牙度过夏天}\\
    It is \{\} a military airfield. & \zh{它\{\}是军用机场}\\
    The restaurant is \{\} the cheapest option available. & \zh{这家餐厅\{\}是最便宜的选择}\\
    The theory is \{\} the only explanation for the phenomenon. & \zh{这个理论\{\}是现象的唯一解释}\\
    This ingredient is \{\} necessary for the recipe. & \zh{这个成分\{\}在食谱中是必要的}\\
    Based on the weather forecast, it will \{\} rain tomorrow. & \zh{根据天气预报，明天\{\}会下雨}\\
    \hline
    \end{tabular}
    }
    \caption{The full list of the English version of Concise Narrative Context (CNC) templates, as well as the corresponding Chinese version. \{\} represents a placeholder for possible WEPs.}\label{sup_cnc_chinese}
\end{table}


By integrating these templates with the seventeen WEPs used in our research, we have compiled a total of 776 prompts for the experiments related to RQ1. We manually adjusted the prompts to ensure grammatical coherence for WEPs like \emph{better than even}, which do not seamlessly fit into the templates. The structured prompting approach enables us to analyze how the inclusion and variation of context influence the LLMs' responses in estimating the certainty associated with different WEPs. 


In addition to the prompts described above, an instruction prompt is appended at the beginning and is constructed as follows: \textit{Format your answer as a float value between 0 and 1, and make your answer short.} Also, to elicit numerical probability estimates from LLMs, we ask the LLM to give its probability estimate using the following template: \textit{Given the statement "\{\}", with what probability do you think \{\}?} The first \{\} is a placeholder for any context template that has been instantiated using a WEP, whereas the second \{\} is a placeholder for the same context template without any instantiation. For example, a fully instantiated prompt, using `Probably' as the WEP and `They will \{\} launch before us' as a CNC context template, would be presented to an LLM as follows: \textit{Format your answer as a float value between 0 and 1, and make your answer short. Given the statement "They will probably launch before us", with what probability do you think they will launch before us?}

We also investigate how a change of prompting language, from English to Chinese, affects the estimative probability for LLMs. Thus, we have curated an additional dataset on the basis of the original English CNC templates (Table \ref{sup_cnc}). We manually translated these CNC templates into Chinese. Similar to the English version, we append the following instruction prompt, translated from the English version with slight variation, at the beginning of any CNC template: \textit{\zh{你的输出只有0到1之间的带有两位小数的浮点值。回答问题时直接给出最终答案，不要加入中间思考过程，不要重复问题。}}. Additionally, to elicit the numerical probability from LLMs, we encapsulate the associated context templates in the following way: \emph{\zh{根据陈述“\{\}”，您认为\{\}的概率是多少？}}, where the first \{\} is a placeholder for any context template that has been instantiated using a WEP and the second \{\} is a placeholder for the same context template without any instantiation. An example using \textit{\zh{可能}} as the WEP and \textit{\zh{他们\{\}在我们之前发布}} as the context template is: \textit{\zh{你的输出只有0到1之间的带有两位小数的浮点值。回答问题时直接给出最终答案，不要加入中间思考过程，不要重复问题。根据陈述“他们很可能会在我们之前发布”，你认为他们在我们之前发布的概率是多少？}}

\subsubsection*{Metrics}
To examine the differences in the underlying distribution of probability estimates of each WEP between humans and LLMs, we utilize both the Kullback-Leibler (KL) divergence test\cite{kullback1951information} and the Mann-Whitney U test.\cite{mann1947test} The KL divergence test provides a measure of how one probability distribution diverges from a second probability distribution. It quantifies the amount of information lost when one distribution is used to approximate another, which is useful in determining how much information is lost when humans interpret probabilities given by LLMs in estimating situations. Mathematically, for discrete distributions, the KL divergence is defined as $D_{KL}(P || Q) = \sum_{i} P(i) \log \left(\frac{P(i)}{Q(i)}\right)$, where $P$ and $Q$ represent the probability distributions. Because the responses gave by humans and LLMs range from 0 to 100, we first discretize the responses by fitting them into 20 bins with equal width, such as 0 to 5, 5 to 10, ..., 90 to 95, and 95 to 100. Then the associated discrete probability distribution can be calculated accordingly, along with KL divergence between any two distributions.  

The Mann-Whitney U test, is a non-parametric statistical test that is used to compare differences between two independent samples. It is useful when the assumptions of normal distribution and homogeneity of variances are not met, as is the case for the human estimations for WEPs. It evaluates whether one group tends to have higher or lower values than the other group, without assuming a specific distribution for the data. Given $n_1$ samples for one population and $n_2$ samples from another, the corresponding Mann–Whitney U statistic is defined as the smaller of: $U_1=n_1n_2+\frac{n_1(n_1+1)}{2}-R_1$, $U_2=n_1n_2+\frac{n_2(n_2+1)}{2}-R_2$. The $R_1$, $R_2$ represent the sum of the ranks in groups 1 and 2, after pooling all samples in one set and where the smallest value obtains rank 1 and so on.


By employing both the KL divergence and Mann-Whitney U test, we gain a more comprehensive understanding of the differences in probability estimates provided by humans and LLMs, highlighting the discrepancies in interpretation and estimation that may exist between these two sources. 

\subsubsection*{Experimental Setup}

Five LLMs are employed in investigating RQ1: GPT-3.5-turbo \cite{GPT-3.5-turbo}, GPT-4 \cite{GPT-4}, LLaMa-7B, LLaMa-13B\cite{touvron2023llama}, and ERNIE-4.0\cite{ERNIE} from Baidu. The first two LLMs are proprietary and significantly larger in scale, whereas the latter two are open-source and comparatively smaller. The last one is an LLM pre-trained primarily using Chinese corpus, whereas the others are pre-trained primarily using English corpus. This setting provides a rich spectrum of comparison points. By analyzing and comparing the responses of these LLMs, we gain insights into the impact of model size and architecture on their differences with humans when interpreting WEPs.
All models, except ERNIE-4.0, have been tested on all four different contexts (CNC, ENC, MCNC, FCNC) and their responses are compared using both metrics with human survey results. Additionally, the GPT family of models and ERNIE-4.0 have been tested on the Chinese version of CNC. For the GPT family of models, all prompts are constructed with the role of `user' without any `system' messages. We used a temperature of 0 to maximize reproducibility for all models.

\subsection*{Investigating the effect of statistical uncertainty on GPT-4's estimative uncertainty}
\subsubsection*{Data Construction}
\begin{figure}[ht]
\centering
\includegraphics[width=\linewidth]{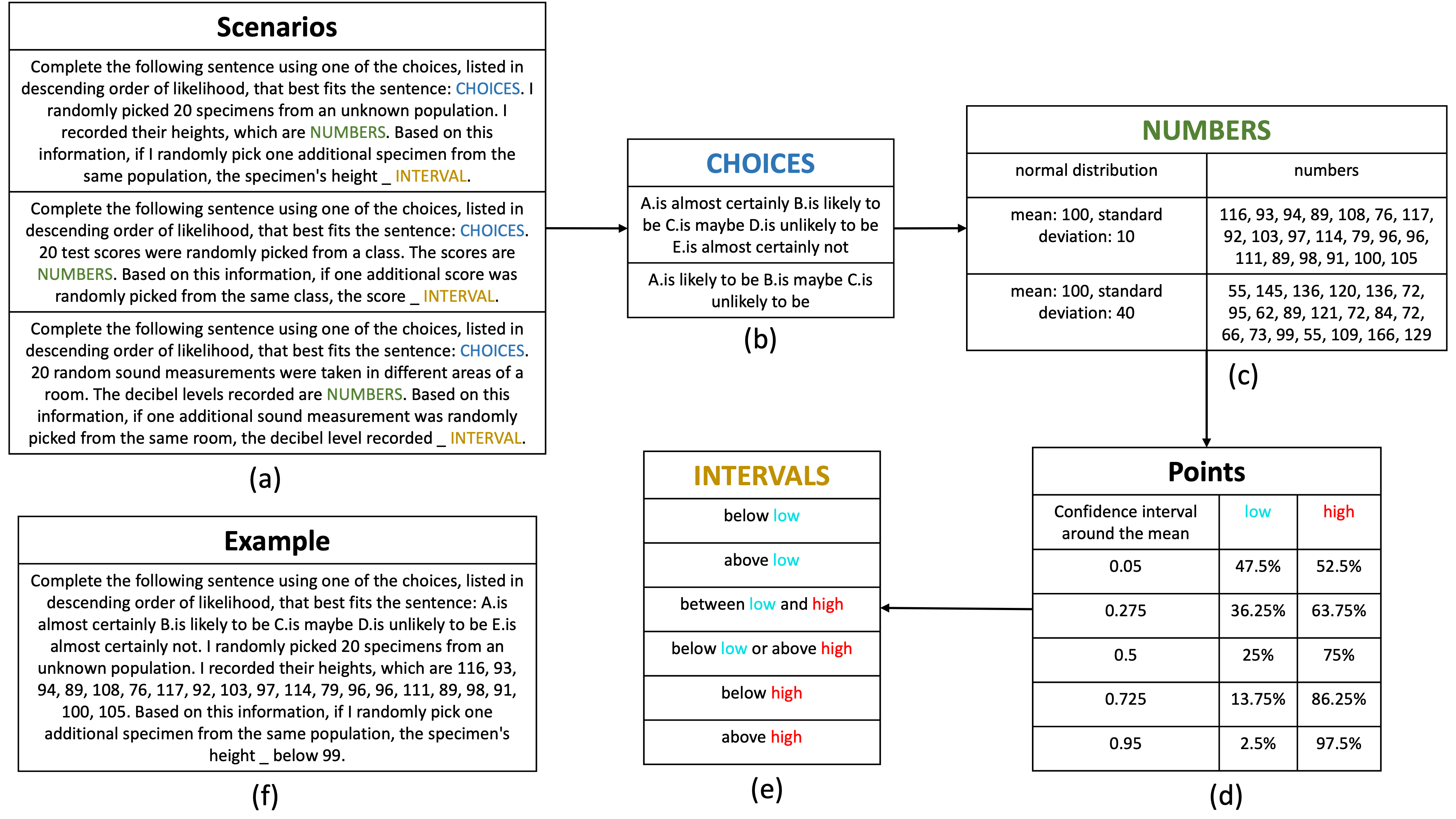}
\caption{The pipeline for constructing the dataset. Three scenario templates (Height, Score, and Sound) are shown in part (a). Each template comes with three controls: CHOICES (b), NUMBERS (c), and INTERVALS (e). Each control offers some possible values. Points (d) contain values that are being used by INTERVALS. One fully constructed example is shown in (f).} 
\label{fig:rq2_method}
\end{figure}

In RQ1, we studied the differences between LLMs and humans in interpreting different WEPs. However, the ability of LLMs to use these WEPs still needs further investigation. In RQ2, we examine how LLMs use the WEPs to express their estimates of statistically uncertain events. Specifically, when given numerical observations of an event's outcome, how would LLMs use a given set of WEPs to estimate the likelihood of future outcomes of the same event? To answer this research question, we constructed the test data set using the pipeline shown in Figure \ref{fig:rq2_method}.

As shown in Figure \ref{fig:rq2_method} (a), three different scenarios (Height, Score, and Sound) were constructed, each with three different controls (CHOICES, NUMBERS, and INTERVAL). Each control represents a distinct variable that influences the outcome in each scenario and is described as follows:

\begin{itemize}
    \item CHOICES: Two sets of possible WEPs choices are available for LLMs to choose from (Figure \ref{fig:rq2_method}(b)). Notice that all of the choices from both sets are used in RQ1. We aim to elicit more fine-grained estimates using the first choice set (i.e., the five choices set) while also investigating the LLMs' estimates under a more generalized scale (i.e., likely, maybe, unlikely) using the other choice set.
    
    \item NUMBERS: Two sets of numbers, one with a narrow range and the other with a wider range (Figure \ref{fig:rq2_method}(c)), can be used to instantiate scenarios. We examine LLMs' estimates under these two different distribution patterns. The first set of numbers is generated based on the normal distribution with a mean of 100 and a standard deviation of 10, whereas the second set is generated based on the normal distribution with a mean of 100 and a standard deviation of 40.

    \item INTERVALS: Six descriptions of mathematical intervals: ($-\infty$, \textit{low}), (\textit{low}, $\infty$), (\textit{low, high}), ($-\infty$, \textit{low}) or (\textit{high}, $\infty$), ($-\infty$, \textit{high}), and (\textit{high}, $\infty$), where \textit{low} and \textit{high} are a pair of integer numbers. When given a set of numbers, we provide LLMs with one of these intervals as a range where these numbers could potentially lie. To generate the pairs of \textit{low} and \textit{high} points, we used the two end points of a confidence interval around the mean of the normal distributions that were used to generate the numbers, where \textit{low} represents the lower end of the interval, and \textit{high} represents the higher end. Five confidence levels were used: 0.05, 0.275, 0.5, 0.725, and 0.95, with each increasing level containing a wider range. The numbers shown in the columns of \textit{low} and \textit{high} in Figure  \ref{fig:rq2_method}(d) represent the probability that an additional number from the same distribution will fall below the point. 
    
    Additionally, three pairs of complementary intervals are defined as: \textit{below low} and \textit{above low}, \emph{between low and high} and \textit{below low or above high}, and \textit{below high} and \textit{above high}. Each of these pairs encompasses the entire range of numbers for any given set of numbers. For example, the interval \textit{below 99} and \textit{above 99} includes all numbers, covering everything less than 99 and everything greater than 99, leaving no number unrepresented except 99 itself.
\end{itemize}

A fully constructed example scenario that we would provide to an LLM is provided in Figure \ref{fig:rq2_method}(f), where the first scenario is instantiated using the \textit{5 choices} control, the normal distribution with mean 100 and standard deviation 10, and the interval \textit{below low} using the confidence interval level at 0.05.

In total, we have three scenarios, two CHOICES sets, two sets of NUMBERS, five confidence interval levels, and six INTERVALS. Hence, we can construct $3\times2\times2\times5\times6=360$ fully instantiated prompts. Additionally, to investigate whether the use of the Chain-of-Thought (CoT) prompting method can bring an increase in performance, a zero-shot CoT\cite{kojima2022large} prompt was generated for each of the 360 constructed prompts. The CoT prompt was generated by changing \textit{Complete the following sentence using one of the choices, listed in descending order of likelihood, that best fits the sentence: CHOICES.} into \textit{First compute the associated probability. Then complete the following sentence using one of the choices, listed in descending order of likelihood, that best fits the sentence: CHOICES. Give your final choice after `I choose:'.}

\subsubsection*{Metrics}

\begin{table}[ht]
\centering
\includegraphics[width=\linewidth]{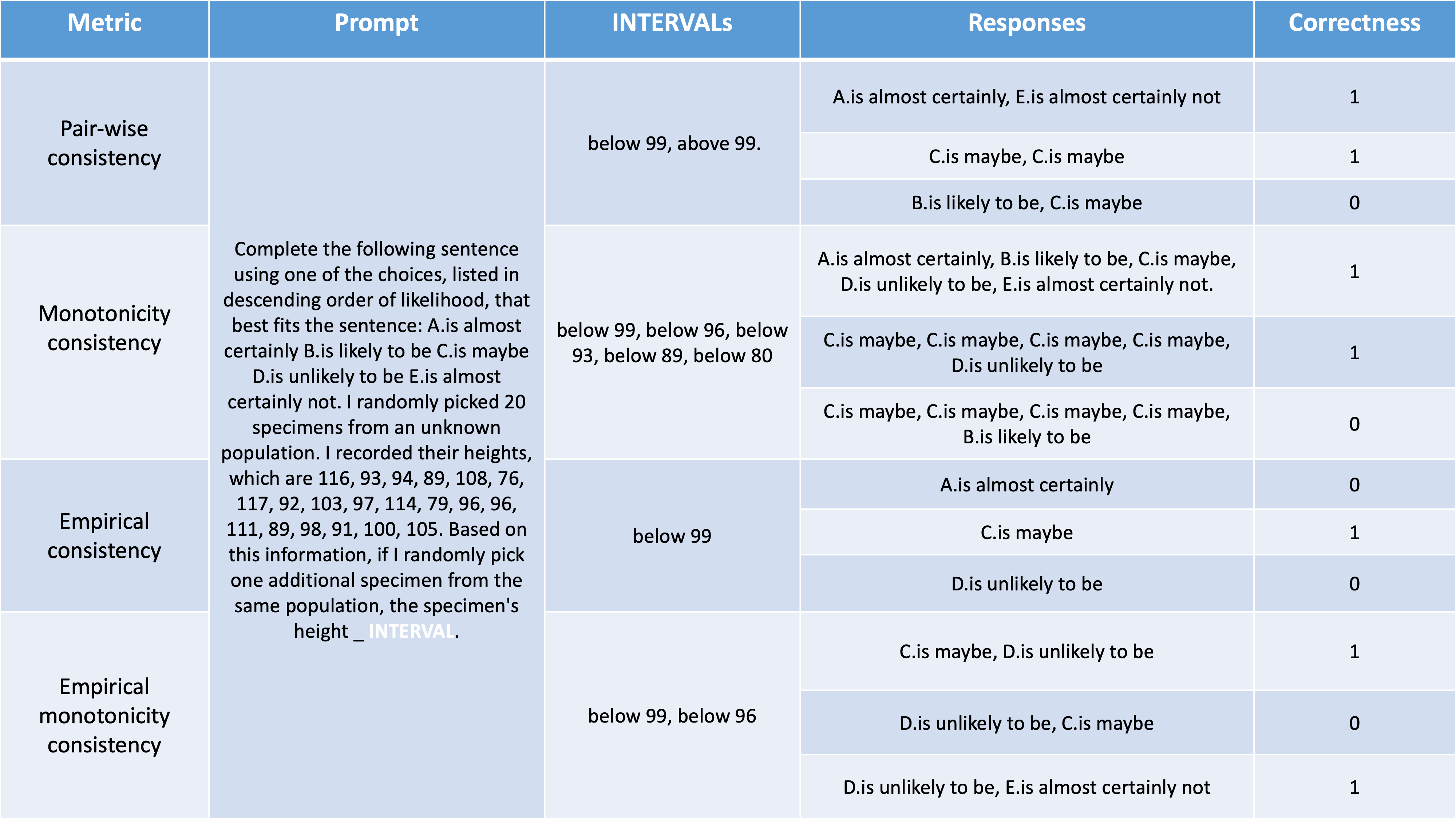}
\caption{Examples showing the correctness of responses (evaluated using the four metrics) when given a prompt that is instantiated with different intervals.} 
\label{fig:rq2_example}
\end{table}

Four metrics were designed to evaluate the model's performances: \textit{pair-wise consistency, monotonicity consistency, empirical consistency}, and \textit{empirical monotonicity consistency}. The model's performance is assessed differently by each of these metrics. These metrics are defined shortly, with an example demonstrated in Table \ref{fig:rq2_example}. We begin by assuming that each prompt $P_{ijks}$ is instantiated using the CHOICES set $C_i$, $i \in \{1,2\}$ the NUMBERS set $N_j$, $j \in \{1,2\}$ and one of the intervals $I_k$, $k \in$ \{\textit{below low}, \textit{above low}, \textit{between low and high}, \textit{below low or above high}, \textit{below high}, \textit{above high}\} from INTERVAL, which is constructed using one of the confidence interval points $P_s$, $s \in \{0.05, 0.275, 0.5, 0.725, 0.95\}$ from \textit{Points}.


\begin{itemize}
    \item \emph{Pair-wise consistency}: Consider two prompts, $P_{ijks}$ and $P_{ijk's}$, that are constructed using the same CHOICES set $C_i$, NUMBERS set $N_j$, and confidence interval point $P_s$, but with different interval such that $k, k' \in$ \{\textit{below low}, \textit{above low}\} or \{\textit{between low and high}, \textit{below low or above high}\} or \{\textit{below high}, \textit{above high}\}. The model's response is deemed correct if and only if it selects any pair of complementary choices for these prompts, regardless of their order. The three pairs of complementary choices are defined as \{\textit{is almost certainly} and \textit{is almost certainly not}\}, \{\textit{is likely to be} and \textit{is unlikely to be}\}, and \{\textit{is maybe} and \textit{is maybe}\}. In total, we have 180 such prompt pairs. Each pair is marked with a 1 if a model answered correctly and a 0 if answered incorrectly. We report the average based on these 180 prompt pairs.

    
    \item \emph{Monotonicity consistency}: For a sequence of five prompts $P_{ijks_1}$, $P_{ijks_2}$, $P_{ijks_3}$, $P_{ijks_4}$, and $P_{ijks_5}$, that are constructed using the same CHOICES set $C_i$, NUMBERS set $N_j$, and interval $I_k$, but with a sequence of increasing confidence interval points, such that $s_1=0.05$, $s_2=0.275$, $s_3=0.5$, $s_4=0.725$, and $s_5=0.95$, the model's response is deemed correct if and only if it selects any sequences of choices that represent probabilities in either an increasing or decreasing order. Specifically, if $k \in$ \{\textit{below low}, \textit{below low or above high}, \textit{above high}\}, the correct order is decreasing, and if $k \in$ \{\textit{above low}, \textit{between low and high}, \textit{below high}\}, the correct order is increasing. Additionally, the rule of increasing or decreasing order is non-exclusive, indicating that the occurrence of two identical WEPs choices does not violate this principle. For example, the sequence of responses (is almost certainly, is almost certainly, is maybe, is maybe, is unlikely to be) counts as a decreasing sequence. In total, we have 72 such sequences of prompts. Each is marked with a 1 if a model answered correctly and a 0 if answered incorrectly. We report the average based on these 72 sequences of prompts.
    
    
    \item \emph{Empirical consistency}: For any prompt $P_{ijks}$, we are able to use the NUMBERS set $N_j$, the interval $I_k$, and the confidence interval point $P_s$ associated with that prompt to calculate the exact proportion of numbers that fall within a specified interval. For example, given the first NUMBERS set (116, 93, 94, 89, 108, 76, 117, 92, 103, 97, 114, 79, 96, 96, 111, 89, 98, 91, 100, 105) and the interval \textit{below 99}, which is \textit{below low} instantiated using the confidence interval point 0.05, there are 12 numbers that fall into the interval. Therefore, the proportion (corresponding to a \emph{frequentist} interpretation of probability) is 0.6. Additionally, for any WEP choice that is provided to GPT-4, we obtained the numerical probability range associated previously with that WEP. Specifically, for the 3-choices CHOICES set, the range for \textit{A.is likely to be} is $(0.61, 1]$, \textit{B.is maybe} is $[0.41, 0.61]$, and \textit{C.is unlikely to be} is $[0, 0.41)$. For the 5-choices CHOICES set, the range for \textit{A.is most certainly} is $(0.92, 1]$, \textit{B.is likely to be} is $(0.61, 0.92]$, \textit{C.is maybe} is $[0.41, 0.61]$, \textit{D.is unlikely to be} is $[0.13, 0.41)$, and \textit{E.is almost certainly not} is $[0, 0.13)$. Based on the calculated proportion and the numerical probability range tied to each WEP choice, we establish the ground truth as the choice whose range encompasses the proportion. In total, we have 360 prompts. Each is marked with a 1 if a model answered correctly and a 0 if answered incorrectly. We report the average based on the 360 prompts.

    
    \item \emph{Empirical monotonicity consistency}: For the sequence of two prompts, $P_{ijks}$ and $P_{ijk's}$, that are constructed using the same CHOICES set $C_i$, NUMBERS set $N_j$, and interval $I_k$, but with a sequence of continuing confidence interval points (i.e., 0.05 and 0.275, 0.275 and 0.5, 0.5 and 0.725, 0.725 and 0.95), the model's response is deemed correct if and only if it selects any sequence of two choices, such that the sequence represents probabilities in an increasing, decreasing, or constant order. This order is determined by first finding out the ground truth for each prompt, which is accomplished in the same way as in the empirical consistency. Then, if the correct choice for the first prompt ($P_{ijks}$) represents a probability greater than that of the second choice ($P_{ijk's}$), the correct order is decreasing. Conversely, if it is lower, the correct order is increasing. If the correct choice for the first and second prompts is the same, the correct order is constant. In total, we have 288 such prompt pairs. Each is marked with a 1 if a model answered correctly and a 0 if answered incorrectly. We report the average based on the 288 prompt pairs.
    
\end{itemize}

\subsubsection*{Experimental Setup}
In the previous experiments, multiple LLMs were studied to investigate the differences in interpreting WEPs between humans and LLMs. However, we only focus on one specific LLM here: GPT-4, as it is among the most powerful LLMs and represents the latest advancements in the field at the time of writing. Similar to the first objective, we used the OpenAI Application Programming Interface (API) to access the GPT-4 model, specifically the `gpt-4-0613' version. All messages sent to the API are constructed without the system message. The prompts are sent only as the role of the `user.' All experiments use a temperature of 0 to maximize reproducibility. All of the 360 normal prompts and 360 CoT prompts are sent to GPT-4 through the official API and responses are recorded. To produce the random performance results for each metric, we randomly choose one choice between the available choices. This process is repeated 10 times, and the final random performance is obtained by averaging the scores for the 10 replications.

\section*{Data availability statement}

All data generated or analyzed during this study are included in this published article [and its supplementary information files].

\section*{Additional information}


\textbf{Accession codes:} Not Applicable; \textbf{Competing interests:} The authors have no competing interests to declare.





\end{document}